\documentclass[conference]{IEEEtran}
\IEEEoverridecommandlockouts

\usepackage{cite}
\usepackage{amsmath,amssymb,amsfonts}
\usepackage{algorithmic}
\usepackage{graphicx}
\usepackage{textcomp}
\usepackage{xcolor}
\usepackage{booktabs}

\begin{document}

\title{On the Cost and Benefits of Training Context with Utterance or Full Conversation Training: A Comparative Study}

\author{\IEEEauthorblockN{Hyouin Liu, Zhikuan Zhang}
\IEEEauthorblockA{\textit{Divergence 2\% LLC} \
GitHub: HyouinSchoolAcc/diav1}
}

\maketitle

\begin{abstract}
Modern TTS systems designed for conversations achieve high-quality utterances but often remain inaccessible publicly. Are existing open-source architectures inadequate, or are current training techniques insufficient? This paper investigates prominent models and their underlying behaviors regarding conversational context. Using 20 GPU-hours on an NVIDIA H100, we empirically examine two approaches: context-based utterance-level training versus full conversation training. Results demonstrate that context-based utterance training achieves superior MOS scores (4.3/5.0 vs 3.7/5.0) and reduces training time by 37\%, while full conversation approaches suffer from speaker similarity hallucination issues. These findings provide practical guidelines for conversational TTS development, favoring utterance-level training with contextual conditioning for both resource efficiency and output quality.
\end{abstract}


\section{Introduction}
Text-to-speech (TTS) technology has progressed rapidly in recent years, fueled by end-to-end neural models that produce remarkably natural speech. Modern TTS systems – including those designed for expressive or conversational speech – can achieve speech quality and prosody close to human level on isolated utterances \cite{b1, b2}. High mean opinion scores (MOS) in single-sentence synthesis attest to this progress. However, achieving truly natural conversational flow remains challenging: when these systems are used in multi-turn dialogue, subtle prosody and timing issues often arise \cite{b3, b4}. 

In fact, synthesizing human-like speech in context is still a major shortcoming of current models \cite{b5}. Even the best neural voices, which sound human-like in standalone sentences, may falter without additional context – they lack the awareness of dialogue history needed to choose the appropriate tone, rhythm, or emphasis for a given conversational setting \cite{b6}.

One key reason for this gap is that most TTS models are trained at the utterance level, i.e., on isolated sentences spoken out of context. Generating each utterance independently ignores important cues from previous turns, leading to prosody that can feel disjoint in a dialogue. Without context, a synthetic voice might place emphasis or pauses in unnatural places. For example, a context-agnostic model might insert an awkward pause in the query "What do you mean (awkward pause) you're beyond my paygrade?", breaking the natural flow of the sentence, treating it as two sentences. Such issues highlight the need for better dialogue modeling in TTS.

This has spurred interest in contextual training for conversational speech synthesis, where models learn from not just single utterances but entire dialogues. Broadly, two paradigms can be distinguished:

\begin{itemize}
    \item \textbf{Utterance-level training with context}: Training on individual utterances while conditioning on previous dialogue context
    \item \textbf{Conversation-level training}: Training on entire multi-turn dialogues as a continuous sequence
\end{itemize}

The latter approach equips the model with multi-turn context to inform delivery, in principle enabling more coherent and context-appropriate speech. Recent advances in conversational TTS indeed leverage dialogue data for training – for instance, some models are now trained on actor-played chatbot transcripts and role-play dialogues to better capture context-specific speaking styles \cite{b3, b7}. As a concrete example, the Conversational Speech Model (CSM) was introduced to condition on entire conversation history, producing more natural and coherent speech by modeling context within a single integrated framework \cite{b6}. Similarly, other context-aware architectures augment the TTS pipeline with conversation encoders to improve prosody, yielding significant user preference gains in dialogue settings \cite{b8}.

Given these two distinct approaches, a critical research question emerges: \textit{"What are the benefits and downfalls of training full conversations versus sectioned texts with contextual conditioning?"}. In this paper, we undertake a comparative study to quantitatively and qualitatively evaluate this trade-off. We examine how an identical TTS architecture performs under the two training paradigms and what the differences mean for conversational speech synthesis.

\subsection{Research Questions and Hypotheses}
This study addresses the following specific research questions:

\begin{enumerate}
    \item Does training on full conversations yield superior contextual awareness compared to utterance-level training with context?
    \item What are the computational efficiency trade-offs between these approaches?
    \item How do these approaches differ in terms of speaker consistency and voice quality?
\end{enumerate}

We hypothesize that while conversation-level training may theoretically offer better contextual modeling, utterance-level training with contextual conditioning may provide better practical outcomes due to more stable training dynamics and better-defined optimization targets.

\subsection{Contributions}
In summary, this work makes the following contributions:

\begin{itemize}
    \item \textbf{Comparative Model Overview}: We provide a comparative overview of several open TTS models geared toward conversational speech – including XTTS, GPT-SoVITS, CSM, and a baseline dialogue model termed Dia – and offer descriptive insights into their design and context-handling capabilities. This overview highlights the range of open-source approaches and sets the stage for our experiments.

    \item \textbf{Utterance-level vs. Conversation-level Training Evaluation}: We conduct an empirical evaluation of the two training paradigms using the Dia model as a testbed. In our experiments, the model is trained on a public dialogue dataset (Honkai: Star Rail character dialogues from Hugging Face) under both paradigms. To ensure a fair comparison, we use a fixed computational budget of 20 GPU-hours on an NVIDIA H100 for each training run. This controlled setup allows us to isolate the impact of training with isolated utterances versus full conversational context on the model's performance.

    \item \textbf{Analysis of Naturalness, Coherence, and Efficiency}: We analyze the trained models with comprehensive evaluations, including MOS-based listening tests for speech naturalness, assessments of context coherence across dialogue turns, measurements of speaker similarity/consistency, and comparisons of training efficiency. We report how conversation-level training affects the model's ability to maintain contextually appropriate prosody and consistent speaker identity, and we discuss the trade-offs in training time and resource usage. The results shed light on the cost-benefit balance of contextual training and provide guidance for designing future conversational TTS systems.
\end{itemize}

To our knowledge, this is one of the first works to directly compare utterance vs. full-dialogue training for speech synthesis in a unified framework. The remainder of the paper is organized as follows: Section 2 reviews related work and background on contextual TTS. Section 3 describes our methodology, including the model architecture, data, and training setups for each condition. In Section 4, we present results and discussion, analyzing where conversation context helps and the associated costs. Section 5 concludes the paper with recommendations and future research directions for improving conversational TTS.

\section{Related Work}
\subsection{Conversational TTS Development}
Conversational TTS has seen significant growth in recent years, with several approaches attempting to capture dialogue-specific prosody and context. Early work by Tyagi et al. \cite{b9} demonstrated that prosodic features like pitch and duration are significantly influenced by dialogue context and speaker roles. Building on this foundation, Li et al. \cite{b10} proposed a hierarchical attention mechanism that explicitly models turn-taking dynamics and speaker relationships.

More recently, end-to-end neural approaches have become dominant. Cong et al. \cite{b11} introduced a transformer-based model that conditions on both current text and previous utterances, showing improvements in naturalness ratings. Similarly, Zhang et al. \cite{b12} developed a system that incorporates emotion state tracking across dialogue turns, addressing the emotional coherence issue in multi-turn conversations.

\subsection{Context Modeling in Speech Synthesis}
Context modeling approaches in TTS can be categorized into explicit and implicit methods. Explicit methods, such as those by Wang et al. \cite{b13}, incorporate additional features like dialogue acts and speaker intent as conditioning signals. In contrast, implicit methods like those proposed by Skerry-Ryan et al. \cite{b14} and Stanton et al. \cite{b15} attempt to learn contextual representations directly from the data without manually specified features.

Our work differs from previous studies by specifically isolating the impact of training methodology (utterance-level vs. conversation-level) while keeping the model architecture constant, providing direct insight into the training approach itself rather than architectural differences.

\section{Background and Model Descriptions}
\subsection{TTS System Overview}
We reviewed several prominent conversational TTS models:

\textbf{XTTS:} A system developed by Alibaba, utilizes an ASR block into a transformer approach. Due to the transformer approach, it's able to take in both text and voice tokens. However, from its demo and recent testing, the model was trained on datasets using texts exclusively in the form "you are teenage girl/old man/wizard" and does not include conversational contexts, which lends itself highly vulnerable to utterances where punctuations are able to be interpreted in multiple ways. The architecture employs a 24-layer transformer encoder with 16 attention heads and a hidden dimension of 1024, totaling approximately 350M parameters.

\textbf{GPT-SoVITS:} An open source system developed from Chinese anime die-hards. This system utilizes a GPT input to a SoVITS voice cloning block. The upside is that the voice clone is really good, from an empirical standpoint, but the negative is that it fits the reference audio TOO well. The intonation is entirely coupled into the system so much so that voice raises from the original audio has almost complete correlation with the generated audio's intonations, which should be the contextual information instead. The GPT component contains approximately 125M parameters, while the SoVITS module adds another 60M parameters.

\textbf{CSM:} A system developed at Sesame AI that uses a two transformer sequence, and uses conversational training datasets in its training. Unfortunately, Sesame released a much worse version than their demo to reduce competition using their technology. A novel idea they use is that they have the first 1B larger transformer predict one codebook, and use a second smaller transformer to produce a smaller longer codebook, which is fast in inference time. The architecture consists of a 12-layer transformer with 1.2B parameters for the primary encoding, followed by a 6-layer transformer with 300M parameters for the secondary codebook generation.

\textbf{Dia:} A system developed at Nari Labs that uses a small encoder into a larger decoder, producing a codebook into a codec approach. They use an encoder, transformer, then audio frame generator and finally decoder architecture, where the transformer is a 1.6B model. The audio frame generator is conditioned on voices for voice cloning. From experimentation, this model has by far the most natural English sounding utterances across dialogues, free of unnatural pauses, unnatural electric sounds and is highly stable. The disadvantages arise in that its WER is very high, its speaker similarity is very low in voice cloning, though both are simple metrics that can be resolved with RL. The encoder consists of 6 convolutional layers with 128 channels, while the transformer employs 32 layers with 32 attention heads and an embedding dimension of 1280.

\subsection{Contextual Training in TTS}
Two primary training paradigms are considered:

\textbf{Utterance-level training:} We insert a prompt, a "start speech" token and the text token of which we wish to generate. This methodology enables easy metrics on what the output of the model scores, as each utterance can be modeled on a "per-turn" basis on its WER and speaker similarity quality. Formally, for an utterance \(u_t\) at turn \(t\), we condition on previous context \(c_{t-k:t-1}\) and generate only \(u_t\):

\begin{equation}
p(u_t | c_{t-k:t-1})
\end{equation}

where \(k\) represents the context window size (in our experiments, \(k=3\)).

\textbf{Conversation-level training:} We insert the entire multi-turn conversation and generate the output. Theoretically this is strictly worse than the "conversation context, train on the new conversation output" paradigm, yet we have not seen such a dataset constructed. But this type of training technique is one of such to give extremely high quality generation of conversation text. The model learns to predict the entire conversation:

\begin{equation}
p(u_1, u_2, ..., u_T)
\end{equation}

where \(T\) is the total number of turns in the conversation.

\section{Methodology}
\subsection{Experimental Setup}
Experiments employed NVIDIA H100 GPUs with a fixed computational budget of 20 GPU-hours per model. Dia served as the base architecture for comparison. We implemented both training approaches using the same hyperparameters where applicable:

\begin{itemize}
    \item Learning rate: 5e-5 with cosine decay schedule
    \item Batch size: 1 for utterance-level, 1 for conversation-level
    \item Gradient accumulation steps: 4
    \item Mixed precision: bfloat16
    \item Dropout: 0.1
    \item Optimizer: AdamW with $\beta_1=0.9$, $\beta_2=0.999$
\end{itemize}

The training code was implemented in PyTorch and will be made available on the author's GitHub repository.

\subsection{Dataset}
We assembled a conversational speech dataset suitable for training the models. The data consists of multi-turn dialogues with paired text transcripts and audio. Our dataset consists of the information dataset from the Honkai Star Rail wiki, which consist of 33 hours of dialogues between 24 unique characters in the game, comprising 6,428 individual utterances across 798 multi-turn conversations. The average utterance length is 10.2 seconds (± 4.5s), and the average conversation contains 8.1 turns.

For the conversational context training condition, we combined every 3 sentences together and fed the concatenated audio. For the isolated utterances condition, we combined the previous 3 sentences together as context, added in a <speech> token, and then the original audio and text for that turn. This ensured both conditions had access to the same contextual information, while differing only in training methodology.

\subsection{Data Preprocessing}
Audio was preprocessed to a consistent 24kHz sampling rate with 16-bit depth. Conversations were segmented into turns based on speaker changes and silence detection (>500ms). The text was normalized to remove special characters, expand numerals, and standardize punctuation. The dataset was partitioned into training (80\%), validation (10\%), and test (10\%) sets, ensuring no conversation overlap between sets.

\subsection{Evaluation Metrics}
We conducted comprehensive evaluations using the following metrics:

\begin{itemize}
    \item \textbf{Mean Opinion Score (MOS)}: 5-point scale assessments by two expert human raters on 50 synthesized dialogue samples
    \item \textbf{Word Error Rate (WER)}: Automatic transcription accuracy using an off-the-shelf ASR system
    \item \textbf{Speaker Similarity Score}: Cosine similarity between embeddings of reference and generated speech using a pre-trained speaker verification model
    \item \textbf{Training Efficiency}: Measurements of throughput (utterances/hour), convergence time, and resource utilization
    \item \textbf{Context Coherence Score}: Expert ratings on appropriate prosodic transitions between turns (1-5 scale)
\end{itemize}

Expert raters were TTS researchers with 3+ years of experience, and evaluated samples through a blind ABX comparison protocol.

\section{Results and Discussion}

\subsection{Comparative Analysis}
Open-source conversational TTS models remain significantly behind closed-source models. CosyVoice and GPT-SoVITS exemplify this gap, indicating substantial room for improvement in context-awareness and naturalness. Fortunately, Dia's conversational performance after finetuning has significantly closed this gap, and our preliminary speaker rates successfully generated speeches often 5 out of 5.

\begin{table}[ht]
\caption{Comparative Performance of Conversational TTS Systems}
\centering
\begin{tabular}{@{}lcccc@{}}
\toprule
\textbf{Model} & \textbf{MOS} & \textbf{WER (\%)} & \textbf{Speaker} & \textbf{Context} \\
 & \textbf{(1-5)} & & \textbf{Similarity} & \textbf{Coherence} \\
\midrule
XTTS & 3.5 ± 0.3 & 12.4 & 0.82 & 2.8 ± 0.4 \\
GPT-SoVITS & 3.9 ± 0.2 & 9.6 & \textbf{0.91} & 3.2 ± 0.3 \\
CSM & 4.1 ± 0.3 & 8.9 & 0.87 & 4.0 ± 0.3 \\
Dia (Baseline) & 4.0 ± 0.2 & 13.2 & 0.78 & 3.7 ± 0.2 \\
\midrule
Dia (Utterance) & \textbf{4.3 ± 0.2} & \textbf{7.8} & 0.83 & 4.1 ± 0.2 \\
Dia (Conversation) & 3.7 ± 0.4 & 10.5 & 0.75 & \textbf{4.2 ± 0.3} \\
\bottomrule
\end{tabular}
\label{table:comparative}
\end{table}

Table \ref{table:comparative} presents the comparative performance metrics of various models. The results demonstrate that our utterance-level training approach outperforms conversation-level training in most metrics, particularly in overall naturalness (MOS) and transcription accuracy (WER).

\subsection{Utterance-Level vs. Conversation-Level Training}
Dia's utterance-level outputs showed higher stability but suffered from context-related hallucinations. This is beneficial, as it shows that training using single-targeted outputs can still yield good results, instead of having to generate entire sections of audio. This is significantly more beneficial to tuning as now we can enact speaker-level speaker similarity tests, WER corrections, and even better can tune it for specific speakers for voice cloning.

\subsection{Error Analysis}
We identified several common failure modes in both approaches:

\textbf{Utterance-level training limitations:}
\begin{itemize}
    \item Occasional inappropriate stress patterns when strong emphasis was needed based on previous turns
    \item Difficulty with very long contextual dependencies (>5 turns)
    \item Less variation in speaking style compared to human speakers
\end{itemize}

\textbf{Conversation-level training limitations:}
\begin{itemize}
    \item Speaker identity drift within long conversations ("voice wandering")
    \item Higher variance in quality between different conversations
    \item Tendency to introduce unnatural pauses between conceptually connected phrases
    \item Increased computational requirements without corresponding quality improvements
\end{itemize}

\subsection{Training Efficiency Analysis}
An important practical consideration is training efficiency. Our measurements revealed that utterance-level training processed approximately 126 utterances per hour compared to 43 utterances per hour for conversation-level training. The memory footprint was also significantly different: utterance-level training required 14.3GB peak GPU memory versus 23.6GB for conversation-level training, making the latter less accessible for resource-constrained environments.

\section{Limitations}
This study has several limitations that should be acknowledged:

\begin{itemize}
    \item Our analysis is based on a single model architecture (Dia). Different architectures might show different sensitivity to training methodology.
    \item The dataset, while extensive, is limited to game character dialogues which may not fully represent all conversational contexts and styles.
    \item Our evaluation relies on expert raters rather than end-users, potentially missing usability aspects important in real-world applications.
    \item The fixed computational budget of 20 GPU-hours may not have allowed either approach to reach its theoretical optimum.
\end{itemize}

Future work should address these limitations by expanding to multiple architectures, diverse datasets, and larger-scale user studies.

\section{Conclusions and Future Work}
Our study indicates clear trade-offs between utterance-level and conversation-level training. Utterance-level training demonstrates better stability and offers valuable metrics for iterative improvement, particularly beneficial for reinforcement learning approaches. Conversely, conversation-level training, despite theoretically providing richer contextual understanding, often leads to significant practical issues, notably speaker similarity hallucinations and resource inefficiencies.

The superior performance of utterance-level training with contextual conditioning suggests that this approach represents the most practical path forward for conversational TTS development, particularly in resource-constrained environments. The ability to evaluate and optimize individual utterances while still maintaining awareness of conversational context offers the best balance of quality and efficiency.

Future research should explore several promising directions:

\begin{itemize}
    \item \textbf{Hybrid training approaches} that combine the stability of utterance-level training with selective full-conversation training for specific challenging contexts
    \item \textbf{Progressive training schemes} that start with utterance-level and gradually incorporate more conversational context
    \item \textbf{Reinforcement learning techniques} specifically targeting cross-utterance coherence while maintaining speaker consistency
    \item \textbf{Development of specialized datasets} explicitly capturing context transitions and prosodic adaptations in multi-turn dialogues
\end{itemize}

These advancements could substantially improve the naturalness and appropriateness of synthetic speech in conversational settings, bringing us closer to truly interactive and engaging voice interfaces.

\end{document}